\documentclass{article} 
\usepackage{iclr2027_conference,times}

\usepackage{amsmath,amsfonts,bm}

\def\eqref#1{equation~\ref{#1}}

\def\1{\bm{1}}

\def\vc{{\bm{c}}}

\def\vv{{\bm{v}}}

\DeclareMathAlphabet{\mathsfit}{\encodingdefault}{\sfdefault}{m}{sl}
\SetMathAlphabet{\mathsfit}{bold}{\encodingdefault}{\sfdefault}{bx}{n}

\usepackage{hyperref}
\usepackage{url}

\usepackage{amssymb}
\usepackage{multirow}
\usepackage{booktabs}

\usepackage{wrapfig}
\usepackage{graphicx}
\usepackage{enumitem}

\usepackage{algorithm}
\usepackage{algpseudocode}
\algrenewcommand\algorithmiccomment[1]{\hfill\small\textcolor{gray}{\textit{// #1}}}

\usepackage{listings}
\usepackage{xcolor}
\title{SGA-Flow-GRPO: Spatial Gradient-Guided Credit Assignment for Flow-GRPO}

\author{
Yunkai Yang$^{1}$ \hspace{0.1em}
Yudong Zhang$^{2}$ \hspace{0.1em}
Xinying Chen$^{3}$ \hspace{0.1em}
Bin Luo$^{4}$ \hspace{0.1em}
Jienan Lyu$^{1}$ \hspace{0.1em}
Kunquan Zhang$^{1}$
\vspace{-2mm}
\AND
\vspace{1mm}
Weitao Wan$^{5}$ \hspace{0.2em}
Runmin Dong$^{1}$\thanks{Corresponding author.}
\\[1mm]
$^{1}$Sun Yat-Sen University \quad
$^{2}$Baidu \quad
$^{3}$Beijing Institute of Technology \\
$^{4}$Tsinghua University \quad
$^{5}$TS Martech \vspace{3mm}
\\
\hspace{20mm}\url{https://github.com/Yunkai-Yang/sga_flow_grpo}
}

\iclrfinalcopy 
\begin{document}

\maketitle

\begin{abstract}
Reinforcement Learning (RL) has proven effective in aligning flow-based generative models with human preferences. Recently, Flow-GRPO has emerged as an efficient critic-free paradigm by calculating advantages over sampled candidate trajectories. However, standard Flow-GRPO applies a uniform scalar advantage across both temporal denoising steps and spatial latent dimensions, without explicitly accounting for the spatial structure of generated images, which may lead to sub-optimal policy updates. To address this, we propose a novel gradient-guided spatial credit assignment framework tailored for Diffusion Transformers (DiTs). We first reformulate the transition-level log-likelihood in Flow-GRPO into a token-wise representation natively aligned with DiT patch architectures, constructing spatially fine-grained importance sampling ratios. To allocate localized credit without rigid, boundary-sensitive segmentation heuristics, we introduce a continuous spatial credit map derived from reward gradients. Crucially, we employ an outlier-robust normalization scheme based on Median Absolute Deviation (MAD) coupled with temperature scaling, effectively eliminating gradient noise while highlighting functional prompt-aligned regions. Extensive evaluations on GenEval show that our approach delivers SOTA alignment quality, achieving a convergence rate comparable to top-tier methods like DiffusionNFT while substantially improving upon Flow-GRPO-based methods in alignment performance.
\end{abstract}

\begin{figure*}[ht]
\centering
\includegraphics[width=1.0\linewidth,keepaspectratio]{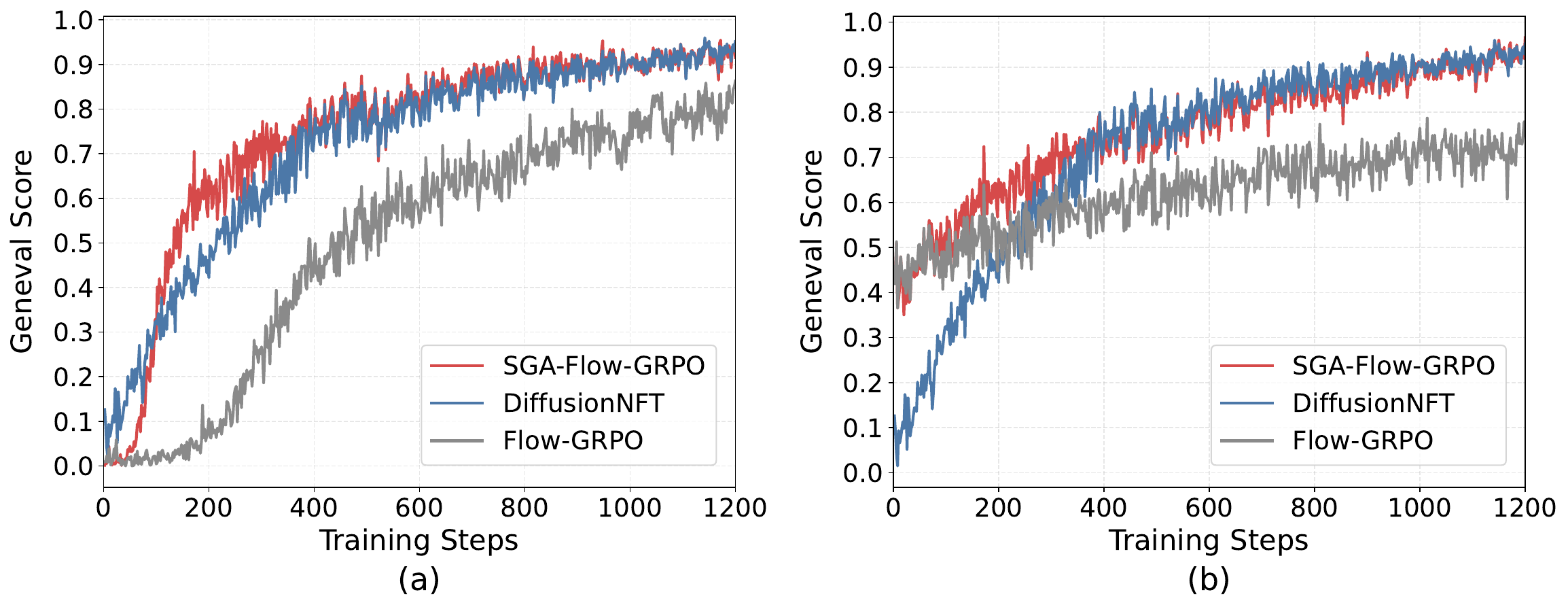}
\vspace{-7mm}
\caption{Training convergence under different CFG scales. (a) $\mathrm{CFG}=1.0$. (b) $\mathrm{CFG}=4.5$. SGA-Flow-GRPO converges comparably to DiffusionNFT and substantially faster than Flow-GRPO.}
\label{fig:teaser}
\vspace{-2mm} 
\end{figure*}

\section{Introduction}
\label{sec:intro}

Diffusion and flow-based generative models have achieved remarkable success in high-fidelity text-to-image synthesis~\citep{ldm_2022, mmdit_2024}. To align generated images with human preferences and fine-grained downstream objectives, Reinforcement Learning (RL) has emerged as a powerful paradigm~\citep{im_video_gen_hf_2026, skyreels_v2_2025}. Existing policy optimization methods generally fall into two main paradigms: critic-based approaches adapting PPO to sequential denoising trajectories~\citep{flow_dppo_2026}, and critic-free methods extending GRPO to flow-based architectures~\citep{flow_grpo_2026}.

While Flow-GRPO provides an efficient and stable alignment framework without an auxiliary critic network, its credit assignment remains coarse: a single scalar advantage is uniformly applied across all denoising steps and spatial latent dimensions. Recent extensions explore temporal credit assignment to differentiate timestep-wise contributions along the denoising trajectory~\citep{stepwise_flow_grpo_2026, tempflow_grpo_2026}, but largely overlook the spatial structure of modern DiT-based models~\citep{dit_2023}. In DiTs, the latent feature map is spatially partitioned into patches and represented as a sequence of tokens, where each token corresponds to a localized region of the generated image. However, Flow-GRPO aggregates the likelihood over all latent dimensions into a single transition-level policy signal, discarding this inherent spatial decomposition. As a result, all spatial tokens receive the same advantage despite contributing differently to the final reward. This mismatch between tokenized generation and uniform credit assignment leads to coarse policy updates and limits optimization efficiency.

A natural way to exploit this spatial structure is to assign differentiated credit to individual latent tokens. One intuitive approach is to identify prompt-relevant regions using off-the-shelf segmenters such as SAM~\citep{sam_2023, sam_v2_2025}. However, hard binary segmentation introduces sharp spatial discontinuities across neighboring latent tokens and lacks fault tolerance-errors in boundary estimation incorrectly penalize critical generative regions and destabilize policy optimization. Conversely, examining the reward gradient backpropagated to the image space reveals a non-trivial challenge: raw gradient magnitude maps are often dominated by unaligned or background regions that absorb extreme gradients to satisfy global loss constraints. Furthermore, naive inversion and Min-Max normalization are extremely vulnerable to gradient outliers, resulting in distorted credit contrast across diverse samples.

Based on these observations, we propose \textbf{SGA-Flow-GRPO}, a spatially fine-grained policy optimization framework tailored to the tokenized structure of DiTs. We first decompose the transition-level log-likelihood of Flow-GRPO into spatially localized token-level components, aligning the policy representation with the patchified latent space. This yields token-wise importance ratios that enable independent optimization across spatial regions. We then derive differentiated token-level advantages from the gradient of the reward with respect to the generated image. To obtain a stable spatial credit map, we normalize reward gradients using Median Absolute Deviation (MAD), followed by a sigmoid transformation, producing continuous regional weights without relying on hard segmentation or manually selected percentile thresholds. The resulting spatial map is aligned with the DiT token layout and used to modulate the group-relative advantage for each token, enabling spatially targeted policy updates. As shown in Fig.~\ref{fig:teaser}, SGA-Flow-GRPO substantially improves the optimization efficiency of Flow-GRPO while achieving state-of-the-art alignment performance on GenEval with convergence speed comparable to DiffusionNFT.

To summarize, our main contributions are summarized as follows:

\begin{itemize}[leftmargin=30pt, itemsep=4pt, topsep=0pt, parsep=1pt]
\item \textbf{Token-Level Likelihood Decomposition:} We reformulate the transition-level log-likelihood in Flow-GRPO into a token-wise representation natively tailored for DiT architectures, constructing fine-grained importance sampling ratios across spatial patch dimensions.

\item \textbf{Reward-Gradient Spatial Credit Assignment:} We propose a continuous, gradient-guided credit allocation mechanism that directly reflects local functional contributions to the global reward, bypassing the rigidity and boundary sensitivity of discrete segmentation masks.

\item \textbf{Superior Performance and Efficiency:} Extensive evaluations on GenEval demonstrate that our method achieves SOTA alignment quality, matching the fast convergence speed of DiffusionNFT while substantially outperforming Flow-GRPO in optimization efficiency.

\end{itemize}
\section{Related Work}
\label{sec:related_work}

\paragraph{RL for Diffusion Models.}

Reinforcement learning has been increasingly explored to align diffusion-based generative models with human preferences and task-specific objectives. Early works formulate diffusion denoising as a sequential decision-making process and optimize the model with reward feedback~\citep{ddpo_2024, dpok_2023}, while subsequent studies explore preference optimization and direct reward backpropagation~\citep{imagereward_2023, hps_v2_2023, diffusion_dpo_2024, aligning_t2i_2023, video_difussion_r_grad}. More recently, group-based policy optimization has been extended to diffusion models. DanceGRPO~\citep{dance_grpo_2025} and Flow-GRPO~\citep{flow_grpo_2026} extend GRPO~\citep{grpo_2024} to diffusion and flow-based models, respectively. Building upon this line, DiffusionNFT~\citep{diffusionnft_2026} and DGPO~\citep{dgpo_2026} further improve the optimization efficiency of flow-based generation, requiring substantially fewer optimization steps than Flow-GRPO. Our method further enhances Flow-GRPO by decomposing the flow model's log-likelihood into token-level components, providing finer-grained optimization signals and enabling more effective credit assignment, which leads to faster convergence than DiffusionNFT.

\paragraph{Credit Assignment for GRPO.}

Credit assignment has been widely studied in LLM reinforcement learning, with recent works exploring token-level credit assignment to provide finer-grained learning signals for language generation~\citep{spo_2026, align_llm_fine_grain_2024, verify_s_by_s_2024, og_a_credit_assign_2026, dense_reward_rlhf_2024}. However, GRPO applies a group-relative advantage uniformly across all tokens, resulting in coarse credit assignment. For diffusion and flow-based generation, this issue is further complicated by the iterative denoising process. Flow-GRPO~\citep{flow_grpo_2026} similarly assigns the final image-level advantage uniformly across all denoising steps. Recent works address this limitation through temporal credit assignment. TempFlow-GRPO~\citep{tempflow_grpo_2026} exploits timestep-dependent importance through trajectory branching and noise-aware weighting, while Stepwise-Flow-GRPO~\citep{stepwise_flow_grpo_2026} assigns credit based on reward improvement at each denoising step. GranularGRPO~\citep{granular_grpo_2026} further explores credit assignment at multiple denoising granularities. Despite these advances, existing methods primarily distinguish contributions across denoising steps, overlooking the spatial structure within generated images.

\section{Methodology}
\label{sec:method}

\subsection{Preliminary}

\paragraph{Policy Optimization.}

Standard policy gradient methods optimize a policy $\pi_\theta$ by maximizing the advantage-weighted log-likelihood of actions. To ensure stable policy updates, PPO~\citep{ppo_2017} replaces the standard objective with a clipped surrogate loss using rollouts from a behavior policy $\pi_{\theta_{\text{old}}}$:
\begin{equation}
\mathcal{L}(\theta) = -\mathbb{E}_{\pi_{\theta_{\text{old}}}} \left[ \min\left( r_t(\theta) \cdot \hat{A}_t, \text{clip}(r_t(\theta), 1-\epsilon, 1+\epsilon) \cdot \hat{A}_t \right) \right],
\end{equation}
where $r_t(\theta) = \frac{\pi_\theta(a_t \mid s_t)}{\pi_{\theta_{\text{old}}}(a_t \mid s_t)}$ is the importance sampling ratio, $\epsilon$ bounds the update step, and $\hat{A}_t$ is the GAE~\citep{gae_2015} computed per action via an auxiliary critic $V_\phi(s)$. For LLM, $s_t$ corresponds to the prompt with context, and $a_t$ represents the generated token at step $t$.

GRPO~\citep{grpo_2024} retains the PPO optimization objective while estimating the advantage without an additional critic network $V_\phi(s)$. Specifically, for diffusion-based generation, given a prompt $c$, GRPO samples a group of $G$ candidate images $\{x^i_0\}_{i=1}^G$ and the corresponding denoising trajectories $\{x_t^i, x_{t-1}^i \dots ,x_0^i\}_{i=1}^G$ from $\pi_{\theta_{\text{old}}}$, assigns the same group-relative advantage to all denoising timesteps within each trajectory:
\begin{equation}
\label{eq:grpo}
\hat{A}_{i}= \frac{R(x^i_0,c) - \text{mean}{(\{R(x^i_0,c)\}_{i=1}^G)}}
{\text{std}({\{R(x^i_0,c)}\}_{i=1}^G)},
\end{equation}
where $R(\cdot)$ denotes the reward function. Furthermore, GRPO introduces a KL divergence loss $D_{KL}(\pi_\theta \parallel \pi_{ref})$ to constrain the policy from drifting too far from the reference model $\pi_{\text{ref}}$.

\paragraph{Flow-GRPO.}

Unlike diffusion model, flow model $v_\theta$ parameterizes a velocity field $v_\theta(x_t,t)$ and generate samples by solving a deterministic ODE. Flow-GRPO~\citep{flow_grpo_2026} converts the ODE into an equivalent SDE that preserves the marginal distribution at each timestep. For Rectified Flow~\citep{rectified_flow_2022}, the updating SDE is
\begin{equation}
x_{t+{\Delta}t} = x_t + \{v_\theta(x_t,t)+\frac{\sigma_t^2}{2t}[x_t+(1-t)v_\theta(x_t,t)]\}\Delta{t}+\sigma_t \sqrt{\Delta{t}} \epsilon
\end{equation}
where $\epsilon \sim \mathcal{N}(0, \boldsymbol{I}) $ and $\sigma_t$ controls the injected stochasticity. The above transition from $x_t$ to $x_{t+\Delta t}$ can be formulated as an isotropic Gaussian policy:
\begin{equation}
\pi_\theta(x_{t+\Delta t}\mid x_t,c) = \mathcal{N} \left (x_{t+\Delta t}; \mu_\theta(x_t,t,c), \sigma_t^2\Delta t \boldsymbol{I}\right),
\end{equation}
where the mean is given by 
$\mu_\theta(x_t,t,c)=x_t+\left\{v_\theta(x_t,t,c)+\frac{\sigma_t^2}{2t}\left[x_t+(1-t)v_\theta(x_t,t,c)\right]\right\}\Delta t.$
Specifically, at each denoising timestep $t$, the importance ratio is given by $
r_t^i(\theta)=\frac{\pi_\theta(x_{t-1}^i \mid x_t^i,c)}{\pi_{\theta_{\mathrm{old}}}(x_{t-1}^i \mid x_t^i,c)}.$
And the KL divergence reduces to
$D_{\mathrm{KL}} \left(\pi_\theta|\pi_{\mathrm{ref}}\right) = \frac{\left( \mu_\theta(x_t,t,c) - \mu_{\mathrm{ref}}(x_t,t,c) \right)^2}{2\sigma_t^2 \Delta t}.$

\subsection{Token-Wise Log-Likelihood}

\begin{figure*}[t]
\centering
\includegraphics[width=1.0\linewidth,keepaspectratio]{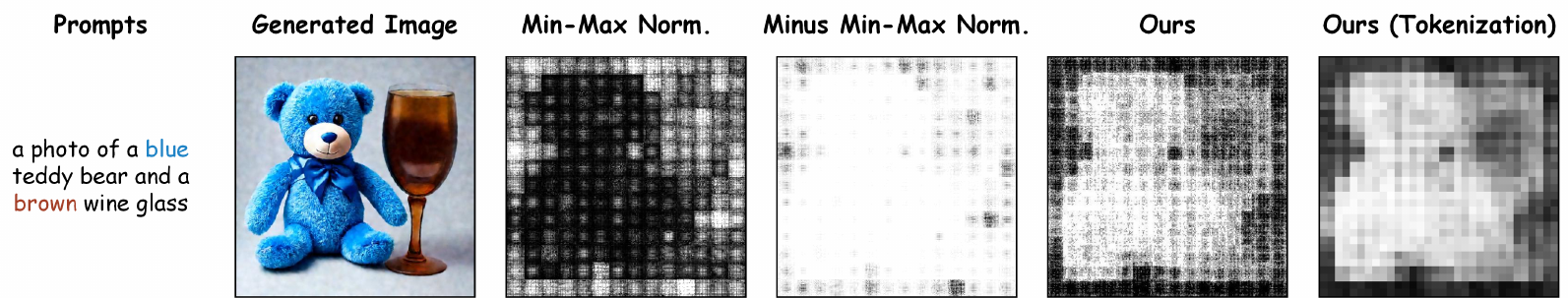}
\vspace{-5mm}
\caption{
Qualitative comparison of spatial credit normalization strategies. Our post-processing better preserves active regions and boundaries than Min-Max normalization (with 0.15 quantile truncation) and its negated variant, yielding spatially consistent credits before and after patchification.
}
\label{fig:grad_map}
\vspace{-2mm} 
\end{figure*}

For each denoising transition from $x_t$ to $x_{t-1}$, standard Flow-GRPO aggregates the log-likelihoods over all latent dimensions into a single transition-level scalar, which is then used to compute the importance ratio $r_t(\theta)$ between the current and old policies. However, in DiT-based flow models~\citep{mmdit_2024, flux_2024}, the latent representation is spatially partitioned into patches and flattened into a sequence of tokens. Each token therefore corresponds to a localized group of latent dimensions, allowing the transition-level log-likelihood to be naturally decomposed into token-level components. Specifically, for a token at spatial position $(X,Y)$, we define its token-level log-likelihood as the sum of the log-likelihoods of the latent dimensions within the corresponding patch:
\begin{equation}
\ell_{(T,X,Y)}(\theta)=
\sum_{(x,y)\in\Omega_{(X,Y)}}
\log \pi_\theta
\left(
x_{T-1,x,y}
\mid
x_{T,x,y},c
\right),
\end{equation}
where $\Omega_{(X,Y)}$ denotes the set of intra-patch spatial coordinates for the token at position $(X,Y)$. The transition-level log-likelihood can thus be decomposed as a sequence of token-level log-likelihoods. Based on this decomposition, we compute a token-wise importance ratio for each token:
\begin{equation}
r_{(T,X,Y)}(\theta)=\frac{\pi_\theta(x_{(T-1, X, Y)} \mid x_{(T, X, Y)}, c)}{\pi_{\theta_{\mathrm{old}}}(x_{(T-1, X, Y)} \mid x_{(T, X, Y)}, c)}
= \exp(\ell_{(T,X,Y)}(\theta) - \ell_{(T,X,Y)}(\theta_{\mathrm{old}})).
\end{equation}
Compared with the transition-level $r(\theta)$, the token-wise ratio provides more diverse signals, enabling finer-grained differentiation of regional contributions for subsequent spatial credit assignment.

\subsection{Reward-Gradient Spatial Credit Assignment}

With the importance ratio $r(\theta)$ tokenized, it is a natural architectural extension to allocate spatially localized advantage values across different tokens, enabling targeted policy optimization. In text-to-image generation, human interest is predominantly concentrated on the specific region designated by the prompt, while the unprompted background plays a secondary role. Consequently, evaluating alignment rewards should naturally reflect this spatial selectivity: scalar rewards typically gauge how faithfully the generated regions adhere to the text conditions. It is thus intuitive to assign fine-grained, localized advantage values primarily to prompt-aligned regions, while downweighting or assigning lower advantages to irrelevant background areas during policy optimization.

A straightforward attempt to isolate prompt-relevant regions is applying zero-shot segmenters such as SAM~\citep{sam_2023}; however, this naive binarization strategy incurs two major failure modes. First, discrete binary masks enforce an abrupt, discontinuous advantage distribution across neighboring latent tokens, unnecessarily zeroing out background credits that remain essential for maintaining overall image fidelity. Second, hard binarization lacks fault tolerance: errors in object boundary estimation result in rigid credit misallocation, where crucial generative tokens may be incorrectly penalized or ignored, ultimately destabilizing policy optimization.

\begin{figure*}[t]
\centering
\includegraphics[width=1.0\linewidth]{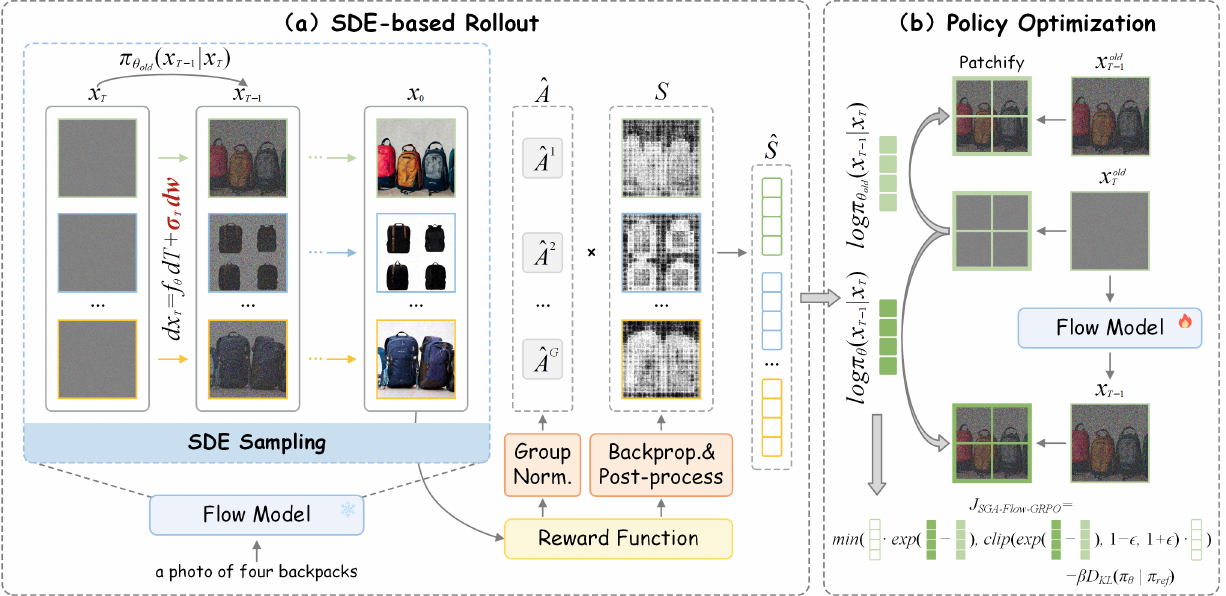}
\vspace{-4mm}
\caption{
Overview of SGA-Flow-GRPO. Given sampled denoising trajectories, our approach decomposes transition-level log-likelihoods into token-wise components to construct fine-grained importance ratios. Meanwhile, reward gradients provide spatial credit signals to modulate the group-relative advantage, enabling differentiated policy updates across spatial tokens.
}
\label{fig:overview}
\vspace{-4mm}
\end{figure*}

To address these challenges, we propose a novel gradient-guided spatial credit assignment mechanism. Instead of relying on rigid, heuristics-based binary segmentation, our method computes spatial credit directly based on the local sensitivity of generated regions with respect to the reward function. By leveraging the reward gradient backpropagated to the latent space, we generate a smooth and continuous spatial credit map where every token position receives a proportional, fine-grained credit allocation corresponding to its functional contribution to the global reward.

Given a prompt $c$, $x_0$ denotes the generated image, $\mathcal{R}(x_0, c)$ is the global scalar reward evaluating the text-conditioned alignment and perceptual quality and $\mathbf{G} \in \mathbb{R}^{h \times w}$ represents the spatial distribution of the reward function's sensitivity with respect to $x_0$:
 
\begin{equation}
\mathbf{G} = \left\| \nabla_{x_0} \mathcal{R}(x_0, c) \right\|_2.
\end{equation}

As shown in Fig.~\ref{fig:grad_map}, directly utilizing the gradient magnitude map $\mathbf{G}$ reveals a counter-intuitive phenomenon: regions with larger gradient norms are often irrelevant to the prompt conditions. Mathematically, the gradient $\nabla_{x_0} \mathcal{R}(x_0, c)$ signifies the local directional change required to elevate the reward. Conversely, unaligned or background regions absorb high-magnitude gradients to satisfy global reward constraints.

However, simply negating $\mathbf{G}$ followed by Min-Max normalization remains severe failure modes. First, the dynamical ranges of gradient norms vary drastically across different generated samples $x_0$. Second, isolated gradient spikes or spatial outliers can distort the entire normalized scale, effectively suppressing the contrast of prompt-relevant regions. Although truncation via fixed percentile clipping can partially mitigate extreme values, the optimal quantile threshold shifts substantially across samples, rendering hard-coded heuristics suboptimal for diverse generation contexts.

To construct a smooth spatial credit map that remains resilient to gradient outliers without requiring manual percentile thresholds, we introduce an outlier-robust normalization scheme based on Median Absolute Deviation (MAD). Unlike standard mean and variance, which are notoriously vulnerable to isolated gradient spikes, median-based statistics provide a high breakdown point that effectively isolates regional anomalies. Specifically, for a gradient norm map $\mathbf{G}$, we first compute the spatial median $\mu_{0.5}(\mathbf{G})$ and the $\mathbf{MAD}(\mathbf{G})$ across all spatial locations of the gradient magnitude map $\mathbf{G}$:

\begin{equation}
\mathbf{MAD}(\mathbf{G}) = \mu_{0.5} \left( \left| \mathbf{G} - \mu_{0.5}(\mathbf{G}) \right| \right).
\end{equation}

Here, $\mathbf{MAD}(\mathbf{G})$ serves as an outlier-resistant scale factor that measures spatial variance without being distorted by extreme gradient magnitudes. With these metrics defined, the gradient magnitude at each pixel location $(x, y)$ is transformed into a standardized robust score:
\begin{equation}
\tilde{\mathbf{G}}_{(x,y)} = \frac{\mathbf{G}_{(x,y)} - \mu_{0.5}(\mathbf{G})}{\mathbf{MAD}(\mathbf{G})},
\end{equation}
which dynamically maps heterogeneous gradient scales across diverse images into a unified, zero-centered metric. Subsequently, to smoothly bound the scores into a continuous credit range, we apply a scaled Sigmoid function $\sigma$ mapping to derive the localized spatial credit map $\mathcal{S}_{x, y} \in (0, 1)$:
\begin{equation}
\label{eq:sigmoid}
\mathcal{S}_{x, y} = \sigma \left( -\frac{\tilde{\mathbf{G}}_{x, y}}{\tau} \right),
\end{equation}
where $\tau > 0$ is a temperature hyperparameter that controls the spatial contrast of credit assignment. Accordingly, to accommodate DiT-based architectures, the spatial credit map $\mathcal{S}$ is patchified to token-level resolution and modulated by the trajectory-level advantage $\hat{A}$, yielding the token-wise advantage $\hat{\mathcal{S}}$ for weighting the corresponding token-level importance ratios. Finally, the overall SGA-Flow-GRPO objective can be formulated as:
\begin{equation}
\label{eq:sga_grpo}
\mathcal{J}_\text{SGA-Flow-GRPO}= \min\left( r^i_t(\theta) \cdot \hat{\mathcal{S}}^i, \text{clip}(r^i_t(\theta), 1-\epsilon, 1+\epsilon) \cdot \hat{\mathcal{S}}^i \right)- \beta \cdot D_{KL}(\pi_\theta \mid \pi_{\text{ref}}) .
\end{equation}
The overview of our method is shown in Fig.~\ref{fig:overview}, while the detailed algorithm is provided in Alg.~\ref{alg:sga_grpo}.

{
\setlength{\textfloatsep}{3pt}
\begin{algorithm}[t]
\caption{SGA-Flow-GRPO Training Process}
\label{alg:sga_grpo}

\textbf{Require:} Pretrained policy $\vv_{\textbf{ref}}$, training policy $\vv_\theta$, reward function $\textbf{R}(\cdot)$, prompt dataset $\mathcal{D}$, Group size $G$, SDE sampling timesteps $T$, Beta $\beta$, Temperature $\tau$, Learning rate $\eta$, EMA decay $\mu$.\\
\textbf{Initialize:} Rollout policy $\vv_{\theta_\text{old}} \leftarrow \vv_\text{ref}$, training policy $\vv_\theta \leftarrow \vv_\text{ref}$.
    
    \begin{algorithmic}[1]
        \For {\text{each iteration} $i$}
            \For {\text{each sampled prompt $\vc \sim \mathcal{D}$}} \Comment{Rollout Step, Trajectory Preparation}
                \State Generate SDE trajectories: $\{x^i_T, x^i_{T-1}, \dots, x^i_0\}_{i=1}^G \sim \vv_{\theta_{\text{old}}}(\cdot \mid \vc)$.
                \State Evaluate $\{x_0^i\}_{i=1}^G$: $\{\textbf{r}^i=\textbf{R}(x_0^i)\}_{i=1}^G$ and compute spatial credit map: $\{\mathcal{S}^i\}_{i=1}^G$.
                \State Compute log-likelihood: $\{\ell_{t}^i=\text{log} \pi_{\theta_\text{old}}(x_t^i \mid x_{t-1}^i)\}^{G, T}_{i=1, t=1}$. \Comment{Shape: [BS, C, H, W]}
                \State Patchify and reduction log-likelihood: $\{\hat{\ell_t^i}=\text{patchify}(\ell_T^i)\}^{G, T}_{i=1, t=1}$. \Comment{Shape: [BS, L]}
            \EndFor
            \State Compute $\{\hat{A}^i\}_{i=1}^G$ using $\{\textbf{r}^i\}_{i=1}^G$ in Eq.~\ref{eq:grpo} for all trajectories.
            \State Assign spatial credit via $\{\hat{A}^i\}_{i=1}^G$: $\{\hat{\mathcal{S}}^i=\hat{A}^i \cdot \mathcal{S}^i\}_{i=1}^G$. \Comment{Shape: [BS, L]}
            \For {each log-likelihood $\hat{\ell_t^i}$} \Comment{Gradient Step, Policy Optimization}
                \State Compute policy log-likelihood: ${\ell_{t}^{i}}' = \text{log} \pi_\theta(x_t^i \mid x_{t-1}^i)$. \Comment{Shape: [BS, C, H, W]}
                \State Patchify and reduction policy log-likelihood: $\hat{\ell_{t}^{i}}'= \text{patchify}({\ell_{t}^{i}}')$. \Comment{Shape: [BS, L]}
                \State Compute important ratio: $r^i_t(\theta)= \text{exp}(\hat{\ell_{t}^{i}}'-\hat{\ell_{t}^i})$ and $D_{KL}(\pi_\theta \mid \pi_{\text{ref}})$.
                \State $\theta \leftarrow \theta - \eta \nabla_{\theta} \left[ \min\left( r^i_t(\theta) \cdot \hat{\mathcal{S}}^i, \text{clip}(r^i_t(\theta), 1-\epsilon, 1+\epsilon) \cdot \hat{\mathcal{S}}^i \right)- \beta D_{KL}(\pi_\theta \mid \pi_{\text{ref}}) \right]$
            \EndFor
            \State Update rollout policy $\vv_{\theta_\text{old}} \leftarrow \mu \cdot \vv_{\theta_\text{old}} + (1-\mu) \cdot \theta$.
        \EndFor
    \end{algorithmic}
\end{algorithm}
}
\section{Experiments}

\begin{table*}[t]
\centering
\caption{Evaluation Results on Geneval.}
\vspace{1mm}
\resizebox{\textwidth}{!}{
    \begin{tabular}{l  c  c  cccccc}
    \toprule
    \textbf{Model} & \textbf{Step}  & \textbf{Overall $\uparrow$ } & \textbf{Single Obj. $\uparrow$} & \textbf{Two Obj. $\uparrow$} & \textbf{Counting $\uparrow$ } & \textbf{Colors $\uparrow$} & \textbf{Position $\uparrow$} & \textbf{Attr. Binding $\uparrow$} \\
    \midrule
    \multicolumn{9}{c}{\textit{Off-the-Shelf Models}} \\
    \midrule
    SD3.5-M (w/o CFG)     &   -  & 0.27 & 0.70 & 0.20 & 0.18 & 0.45 & 0.04 & 0.07 \\
    SD3.5-M               &   -  & 0.63 & 0.98 & 0.78 & 0.50 & 0.81 & 0.24 & 0.52 \\
    FLUX.1 Dev            &   -  & 0.66 & 0.98 & 0.81 & 0.74 & 0.79 & 0.22 & 0.45 \\
    GPT-4o                &   -  & 0.84 & 0.99 & 0.92 & 0.85 & 0.92 & 0.75 & 0.61 \\
    \midrule
    \multicolumn{9}{c}{\textit{Post-training Methods}} \\
    \midrule
    Flow-GRPO (w/o CFG)   & 2.4k & 0.84 & 0.99 & 0.97 & 0.86 & 0.83 & 0.67 & 0.74 \\
    Flow-GRPO             &  2k  & 0.69 & 0.98 & 0.83 & 0.56 & 0.60 & 0.57 & 0.60 \\
                          & 4.8k & 0.92 & \textbf{1.00} & 0.98 & 0.93 & 0.89 & 0.85 & 0.87 \\
    DGPO                  &  2k  & 0.93 & 0.99 & \textbf{0.99} & 0.94 & 0.90 & 0.87 & 0.88 \\
    TempFlow-GRPO         &  3k  & 0.94 & \textbf{1.00} & 0.98 & \textbf{0.96} & \textbf{0.92} & 0.87 & 0.88 \\
    DiffusionNFT          & 1.7k & 0.94 & \textbf{1.00} & \textbf{0.99} & \textbf{0.96} & 0.91 & 0.86 & 0.90 \\
    Ours (w/o CFG)        & 1.8k & 0.94 & \textbf{1.00} & \textbf{0.99} & 0.95 & 0.91 & 0.89 & 0.87 \\
    Ours                  & 1.8k & \textbf{0.95} & \textbf{1.00} & \textbf{0.99} & \textbf{0.96} & \textbf{0.92} & \textbf{0.91} & \textbf{0.91} \\
    \bottomrule
    \end{tabular}
}
\label{tab:geneval}
\end{table*}

\subsection{Experimental Setup}

\paragraph{Implementation Details.}

We fine-tune SD3.5-Medium~\citep{mmdit_2024} at a resolution of $512 \times 512$ using LoRA with $\alpha=64$ and $r=32$. Training is conducted on 8 NVIDIA A100 GPUs using an implementation built upon the Flow-GRPO codebase. For policy training, we utilize an SDE solver with $10$ timesteps for trajectory sampling, with the KL divergence weight set to $0.004$. The training batch configuration consists of a group size of $24$ with $48$ groups per epoch. Experiments are conducted separately for Classifier-Free Guidance (CFG)~\citep{cfg_2022} scales of $1.0$ (without guidance) and $4.5$. During evaluation, trajectory resolution is increased to $40$ timesteps using an ODE solver. All experiments were conducted on GenEval~\citep{geneval_2023}.

\paragraph{Reward Function.}

We use different reward functions under different CFG settings. For $\mathrm{CFG}=1.0$, we combine GenEval Score~\citep{geneval_2023}, ImageReward~\citep{imagereward_2023}, and PickScore~\citep{pickscore_2023}. For $\mathrm{CFG}=4.5$, we use GenEval alone, as detailed in Sec.~\ref{sec:ab_study_1}, where additional rewards yield only marginal convergence gains. Unless otherwise specified, all baseline comparisons use only the GenEval Score.

\begin{figure*}[ht]
\centering
\includegraphics[width=0.95\linewidth]{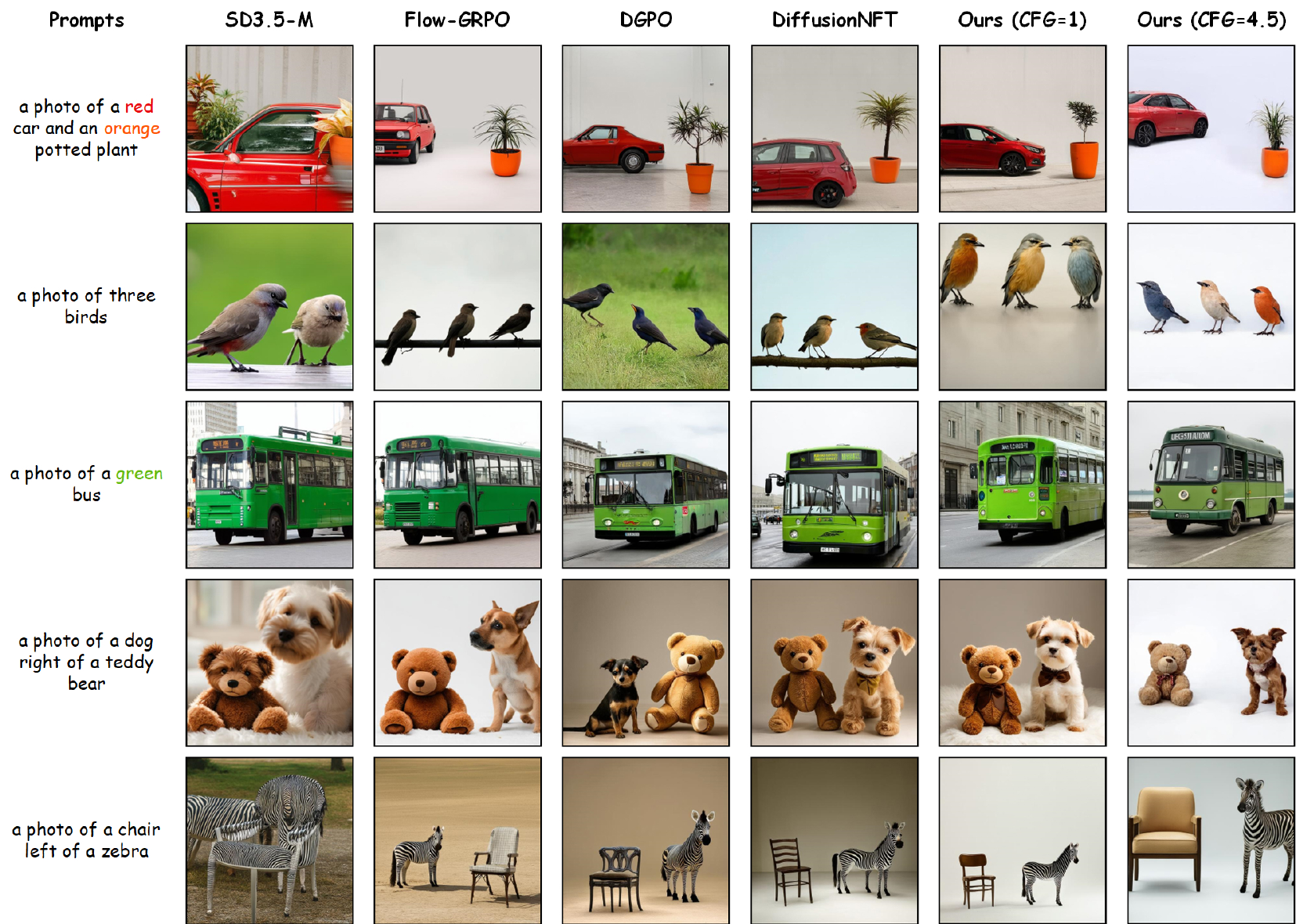}
\caption{Qualitative Comparison on the GenEval Benchmark.}
\label{fig:comparison}
\end{figure*}

\paragraph{Differentiable Reward Function.}

Our gradient-guided spatial credit assignment requires a differentiable reward function to backpropagate gradients to the generated image and construct spatial credit maps. However, the GenEval score is computed using non-differentiable evaluation operators and therefore cannot directly provide the gradient signal required by our method. We thus employ PickScore and ImageReward as differentiable auxiliary reward models for gradient estimation.

Under $\mathrm{CFG}=1.0$, following the multi-reward configuration described above, GenEval, PickScore, and ImageReward jointly contribute to the scalar reward used for advantage estimation, while the spatial gradients are computed from the differentiable PickScore and ImageReward components. Under $\mathrm{CFG}=4.5$, the scalar reward is restricted to GenEval alone to ensure a fair comparison with Flow-GRPO and other baselines. In this case, PickScore and ImageReward are used exclusively to provide auxiliary gradients for spatial credit estimation and do not contribute to either the scalar reward or the group-relative advantage. This design preserves the same optimization reward as the baselines while introducing differentiable supervision only for spatial credit assignment.

\subsection{Image Generation Comparison}

As summarized in Table~\ref{tab:geneval}, we evaluate our approach against off-the-shelf models and leading post-training alignment baselines on the GenEval benchmark, including SD3.5-M~\citep{mmdit_2024}, FLUX.1 Dev~\citep{flux_2024}, GPT-4o~\citep{gpt_4o_2024}, Flow-GRPO~\citep{flow_grpo_2026}, DGPO~\citep{dgpo_2026}, TempFlow-GRPO~\citep{tempflow_grpo_2026}, DiffusionNFT~\citep{diffusionnft_2026}. All post-training methods use the same group size, groups per epoch, and two update steps per epoch as our method.

Compared with Flow-GRPO-based methods, our approach provides a substantial improvement in optimization efficiency. Flow-GRPO reaches an overall score of 0.92 after 4.8k training steps, while TempFlow-GRPO achieves 0.94 after 3k steps. In contrast, our method reaches 0.95 with only 1.8k steps and performs particularly well on challenging relational categories such as \textit{Position} and \textit{Attribute Binding}. These results show that our method achieves stronger compositional alignment with substantially fewer optimization steps than Flow-GRPO-based approaches.

We further compare our method with DGPO and DiffusionNFT, which follow a forward-optimization paradigm. DiffusionNFT reaches 0.94 after 1.7k steps, while our method achieves 0.95 with a comparable budget of 1.8k steps. As shown in Fig.~\ref{fig:comparison}, our method also exhibits stronger qualitative alignment, particularly in spatial relationships and attribute correspondences, consistent with its gains on \textit{Position} and \textit{Attribute Binding}, indicating that our method combines fast convergence with improved fine-grained compositional alignment.

Furthermore, our method remains consistent across different CFG settings. With $\mathrm{CFG}=4.5$, the overall score improves from 0.63 to 0.95, while under $\mathrm{CFG}=1.0$, it increases from 0.27 to 0.94. The small performance gap suggests that the proposed optimization is robust to different guidance strengths and generalizes well across CFG configurations.

\subsection{Ablation Studies}
\label{sec:ab_study}

\paragraph{Ablation of Log-likelihood Tokenization.}

\begin{figure*}[h]
\centering
\includegraphics[width=1.0\linewidth,keepaspectratio]{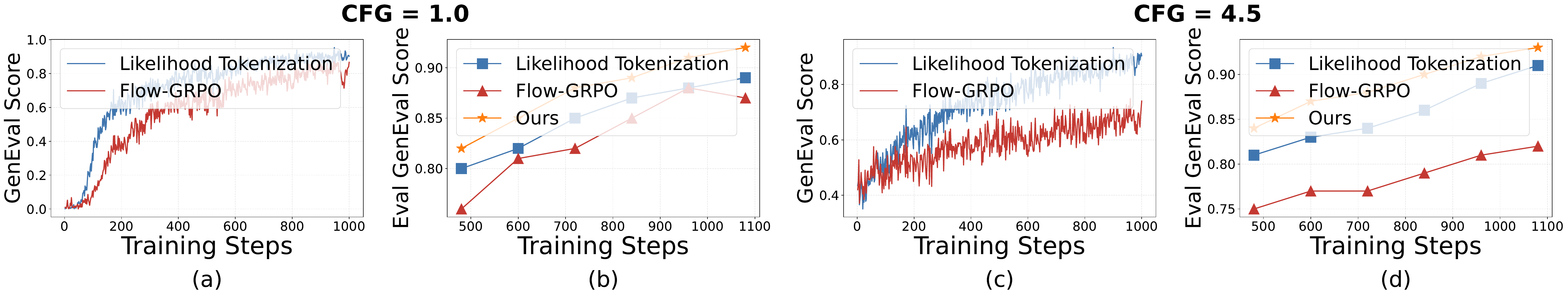}
\vspace{-5mm}
\caption{Ablation of log-likelihood tokenization and spatial credit assignment. Under $\mathrm{CFG}=1.0$ and $\mathrm{CFG}=4.5$, we report the training convergence curves and the GenEval scores on the test set.}
 \label{fig:log_tokenization}
 \vspace{-2mm}
\end{figure*}

We first investigate the contribution of log-likelihood tokenization to optimization efficiency. As shown in Fig.~\ref{fig:log_tokenization} (a)(c), replacing the transition-level likelihood with token-wise log-likelihood consistently accelerates training under both $\mathrm{CFG}=1.0$ and $\mathrm{CFG}=4.5$. Compared with Flow-GRPO, the tokenized formulation converges substantially faster, indicating that decomposing the image-level policy signal into spatially localized token-wise ratios provides a more effective optimization signal.

We further evaluate the corresponding GenEval performance on the test set. Under $\mathrm{CFG}=1.0$ and $\mathrm{CFG}=4.5$, log-likelihood tokenization consistently improves over Flow-GRPO, while our full method achieves the best performance throughout training. Notably, adding spatial credit assignment on top of log-likelihood tokenization does not markedly alter the overall convergence speed, but further improves the evaluation score under a comparable training budget. These results suggest that log-likelihood tokenization primarily drives the gain in optimization efficiency, whereas spatial credit assignment enables finer-grained policy updates and stronger alignment.

\paragraph{Ablation of Reward Function Combination.}
\label{sec:ab_study_1}

\begin{wrapfigure}{r}{0.40\columnwidth}
    \centering
    \includegraphics[width=0.38\columnwidth]{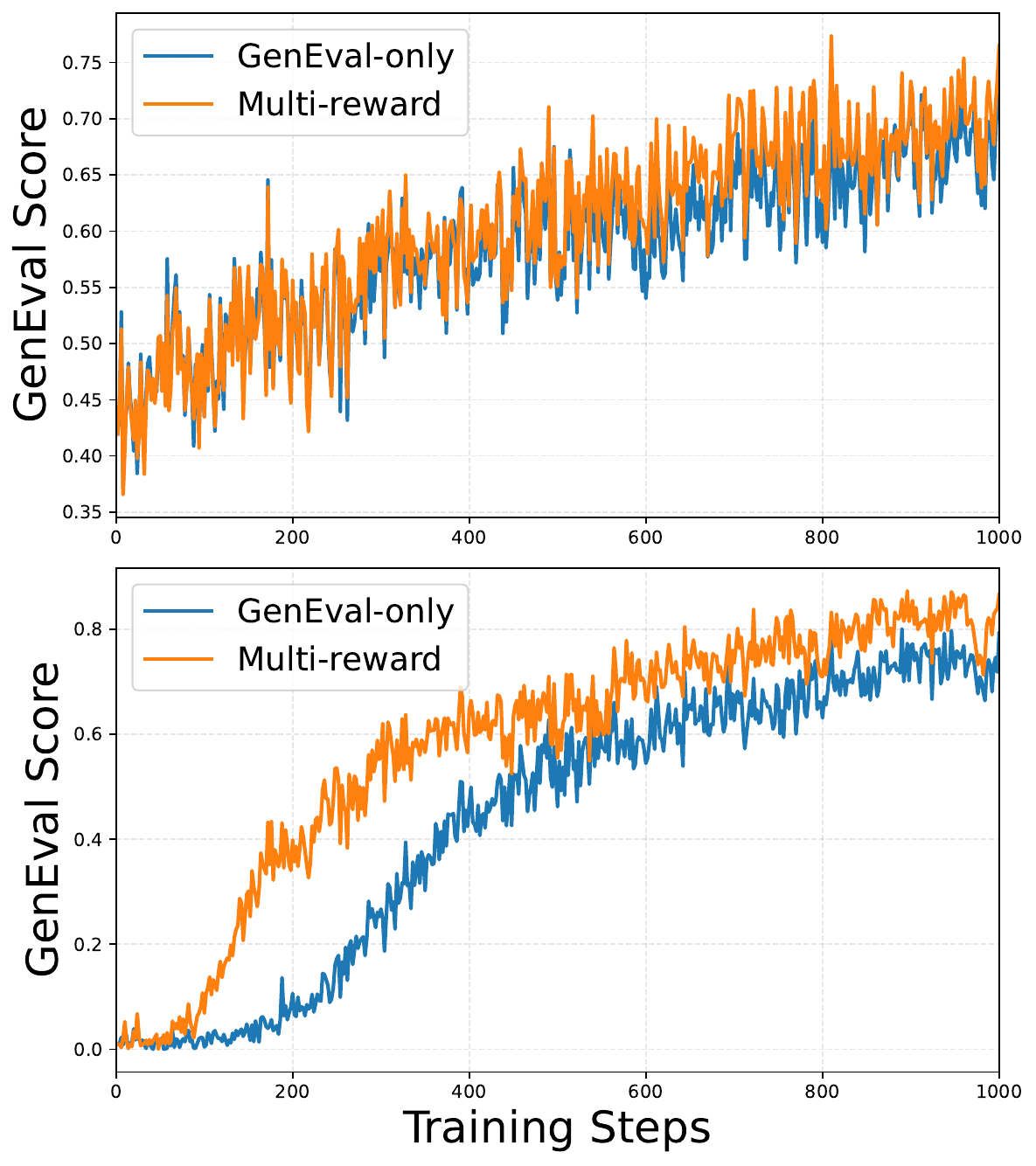}
    \caption{Performance comparison between Multi-reward and GenEval-only optimization under $\mathrm{CFG}=4.5$ (top) and $\mathrm{CFG}=1.0$ (bottom).}
    \label{fig:multi_rewards}
    \vspace{-5mm}
\end{wrapfigure}

We investigate the effect of reward combinations under different CFG scales. As shown in Figure~\ref{fig:multi_rewards}, multi-reward optimization with GenEval, ImageReward, and PickScore substantially accelerates early-stage convergence when $\mathrm{CFG}=1.0$, whereas its advantage becomes marginal under $\mathrm{CFG}=4.5$.

This difference mainly arises from reward diversity during early training. Under $\mathrm{CFG}=1.0$, the stochasticity introduced by the SDE formulation in Flow-GRPO can lead to low-quality samples at the beginning of training, resulting in consistently low GenEval scores during the early stage of training. Consequently, the GenEval-only setting provides low-variance group rewards and therefore weak relative advantages for GRPO optimization. By incorporating additional learned image-text preference signals, the multi-reward setting can still distinguish samples even when their GenEval scores are uniformly low, thereby yielding more informative optimization signals. In contrast, at $\mathrm{CFG}=4.5$, GenEval alone already provides sufficient reward variation, leaving limited room for additional rewards to improve convergence.

\paragraph{Ablation of Temperature in Reward-Gradient Map.}

\begin{wrapfigure}{r}{0.64\columnwidth}
    \centering
    \includegraphics[width=0.62\columnwidth]{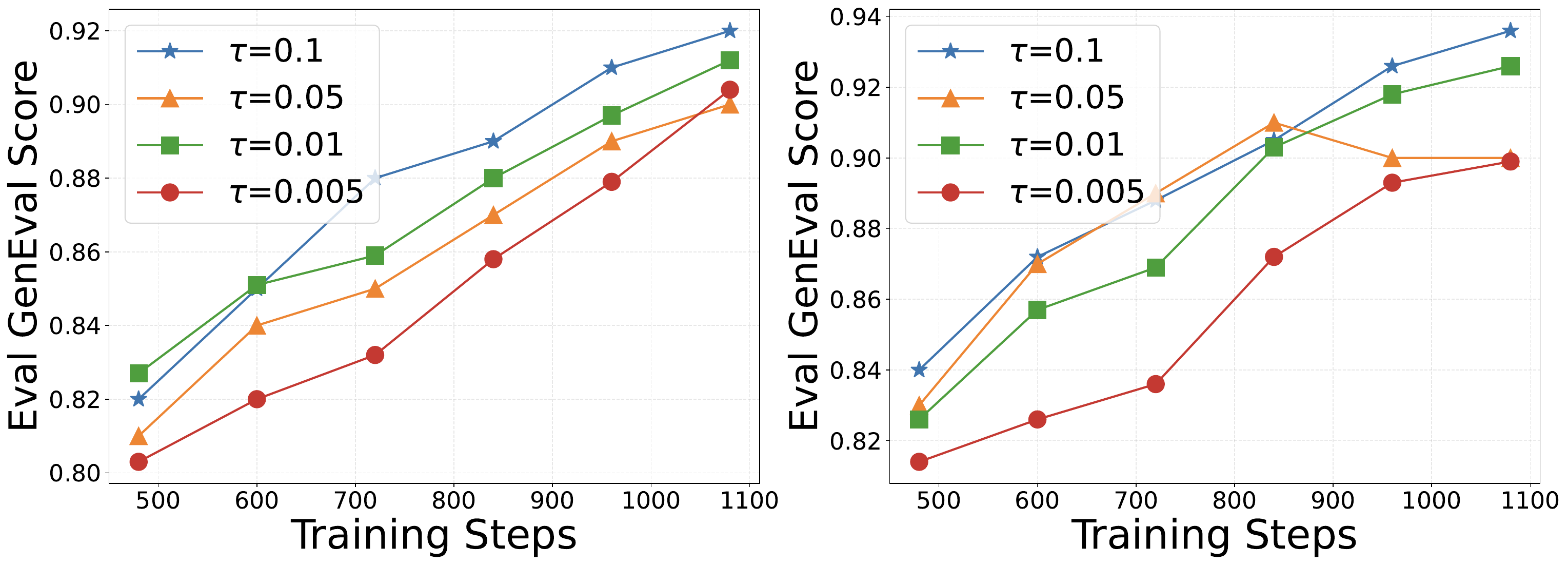}
    \vspace{-2mm}
    \caption{Ablation of temperature $\tau$ in the reward-gradient spatial map under $\mathrm{CFG}=1.0$ (left) and $\mathrm{CFG}=4.5$ (right), evaluated by GenEval score on the test set.}
    \label{fig:tau}
    \vspace{-2mm}
\end{wrapfigure}

We further study the effect of the temperature parameter $\tau$ in the reward-gradient spatial map. As defined in Eq.~\ref{eq:sigmoid}, $\tau$ controls the sensitivity of the Sigmoid mapping to the normalized reward gradient and thus determines the spatial contrast of the resulting credit map. A smaller $\tau$ produces sharper spatial responses and increases sensitivity to local gradient variations, which can improve the localization of credit assignment but may also amplify noisy or unstable gradients. In contrast, a larger $\tau$ yields smoother credit maps with less aggressive regional separation, providing more stable spatial weighting at the cost of reduced local sensitivity.

As shown in Fig.~\ref{fig:tau}, we compare $\tau \in \{0.1,\, 0.05,\, 0.01,\, 0.005\}$ under both $\mathrm{CFG}=1.0$ and $\mathrm{CFG}=4.5$. Different values of $\tau$ exhibit similar convergence speeds, indicating that the temperature mainly affects the quality of spatial credit allocation rather than optimization efficiency. In both settings, $\tau=0.1$ achieves the best GenEval performance on the test set. This suggests that, within the evaluated range, a relatively smoother reward-gradient map provides a better balance between spatial discrimination and optimization stability, whereas overly small temperatures may produce excessively sharp credit allocation and reduce the robustness of policy updates.
\section{Conclusion}

We introduce SGA-Flow-GRPO, a spatially fine-grained credit assignment framework for reinforcement learning of flow-based generative models. Our approach decomposes the transition-level likelihood in Flow-GRPO into token-wise components, enabling policy optimization at the spatial token level, and constructs a reward-gradient spatial credit map to assign differentiated advantages across image regions. To improve the robustness of gradient-based credit estimation, we employ MAD-based normalization with temperature scaling, avoiding the brittleness of hard segmentation and outlier-sensitive normalization. Experiments on GenEval demonstrate that token-wise likelihood modeling substantially improves optimization efficiency, while spatial credit assignment enhances fine-grained compositional alignment. Overall, our results highlight the importance of spatially structured credit assignment for efficient reinforcement learning of DiT-based flow models and suggest a promising direction toward more precise, localized policy optimization for generative models.

\bibliography{iclr2027_conference}
\bibliographystyle{iclr2027_conference}

\appendix
\clearpage
\newpage

\section{The Use of Large Language Models (LLMS)}

We used large language models (LLMs) solely as a writing assistant for language polishing and improving clarity of presentation. The LLMs were not involved in research ideation, methodological design, experimental execution, or result analysis. All scientific contributions and substantive writing were carried out by the authors.

\section{Additional Algorithmic Details}

\subsection{Log-Likelihood Estimation under SDE Sampling}

For a general flow ODE, the corresponding reverse-time SDE can be written as
\begin{equation}
d x_t=\left[ v_\theta(x_t,t,c) - \frac{\sigma_t^2}{2} \nabla_{x_t}\log p_t(x_t) \right]dt + \sigma_t\,d w_t,
\label{eq:sde_general}
\end{equation}
where $\sigma_t$ controls the injected stochasticity and $w_t$ denotes a
standard Wiener process. For Rectified Flow, whose interpolation is defined
as
\begin{equation}
    x_t = (1-t)x_0 + t x_1,
\end{equation}
the score function can be expressed through the velocity field, leading to
the SDE used in Flow-GRPO:
\begin{equation}
d x_t = \left[ v_\theta(x_t,t,c) + \frac{\sigma_t^2}{2t} \left( x_t + (1-t)v_\theta(x_t,t,c) \right) \right]dt + \sigma_t\,d w_t.
\label{eq:rectified_flow_sde}
\end{equation}

Applying Euler-Maruyama discretization with step size $\Delta t$ gives
\begin{equation}
\begin{aligned}
x_{t+\Delta t} = x_t + \left[ v_\theta(x_t,t,c) + \frac{\sigma_t^2}{2t} \left( x_t + (1-t)v_\theta(x_t,t,c) \right) \right]\Delta t + \sigma_t\sqrt{\Delta t}\,\epsilon, \qquad \epsilon\sim\mathcal{N}(0,I).
\end{aligned}
\label{eq:em_discretization}
\end{equation}

Importantly, Eq.~\eqref{eq:em_discretization} induces an explicit Gaussian
transition distribution. Defining
\begin{equation}
\begin{aligned}
\mu_\theta(x_t,t,c) = x_t + \left[ v_\theta(x_t,t,c) + \frac{\sigma_t^2}{2t} \left( x_t + (1-t)v_\theta(x_t,t,c) \right) \right] \Delta t,
\end{aligned}
\label{eq:sde_mean}
\end{equation}
the stochastic transition can be interpreted as the policy
\begin{equation}
\pi_\theta(x_{t+\Delta t}\mid x_t,c) = \mathcal{N} \left( x_{t+\Delta t}; \mu_\theta(x_t,t,c), \sigma_t^2\Delta t\,I \right).
\label{eq:sde_policy}
\end{equation}

Therefore, the SDE transition defines an element-wise Gaussian policy.
Let $h=-\Delta t>0$ denote the reverse-time step size. For each latent
dimension $d$, we have
\begin{equation}
x_{t-h}^{(d)} \sim \mathcal{N} \left( \mu_{\theta}^{(d)}(x_t,t,c), \sigma_t^2 h \right).
\end{equation}
Accordingly, its probability density is
\begin{equation}
\pi_\theta^{(d)} = \frac{1}{\sqrt{2\pi}\sigma_t\sqrt{h}} \exp \left[ -\frac{ \left( x_{t-h}^{(d)} - \mu_\theta^{(d)} \right)^2}{2\sigma_t^2 h}\right].
\end{equation}

Taking the logarithm gives the log-likelihood of each latent element:
\begin{equation}
\boxed{
\log \pi_\theta^{(d)} = - \frac{ \left( x_{t-h}^{(d)} - \mu_\theta^{(d)} \right)^2}{2\sigma_t^2 h } - \log\left(\sigma_t\sqrt{h}\right) - \frac{1}{2}\log(2\pi)}.
\label{eq:element_log_prob}
\end{equation}

For the transition-level log-likelihood, the element-wise log-likelihoods are simply summed over all latent dimensions, yielding a single scalar tensor for each transition. In contrast, for the token-level log-likelihood, we only aggregate the elements within each spatial patch, resulting in $L$ token-wise log-likelihood tensors, where $L$ denotes the number of latent tokens.

\section{Additional Implementation Details}

\subsection{Log-Likelihood Tokenization}

\begin{algorithm}[t]
\caption{Token-Level Log-Likelihood Computation}
\label{alg:log_likelihood}
\begin{algorithmic}[1]
\Require Velocity Prediction $v_\theta(x_t,t)$, Noise Scale $\sigma_t$, Reverse Step Size $h=-\Delta t>0$, Latent State $x_t$, Gaussian Noise $\epsilon\sim\mathcal{N}(0,I)$, and Timestep $t$

\State Compute the transition mean:
\[ \mu_t = x_t \left( 1-\frac{\sigma_t^2}{2t}h \right) - v_\theta(x_t,t) \left( 1+\frac{\sigma_t^2(1-t)}{2t} \right)h. \]

\State Sample the next latent state:
\[ x_{t+\Delta t} = \mu_t + \sigma_t\sqrt{h}\,\epsilon. \]

\State Compute the element-wise log-likelihood:
\[ \ell = -\frac{ \left(x_{t+\Delta t}-\mu_t\right)^2 }{ 2\sigma_t^2 h } -\log\left(\sigma_t\sqrt{h}\right) -\frac{1}{2}\log(2\pi). \]

\State Patchify the element-wise log-likelihood:
\[ \hat{\ell} = \operatorname{Patchify}(\ell). \] \Comment{$[B,C,H,W]\rightarrow[B,L,D_p]$}

\State Compute the token-level log-likelihood:
\[
\ell_{\mathrm{token},l} = \sum_{d=1}^{D_p} \hat{\ell}_{l,d}, \qquad l \in \{1,\ldots,L\}. \] \Comment{$[B,L,D_p]\rightarrow[B,L]$}

\end{algorithmic}
\end{algorithm}

Alg.~\ref{alg:log_likelihood} provides a more detailed implementation of the log-likelihood computation under SDE sampling. Specifically, the Euler-Maruyama discretization induces a Gaussian transition distribution, from which the element-wise log-likelihood can be evaluated in closed form. Standard Flow-GRPO aggregates these element-wise log-likelihoods over all latent dimensions, resulting in a single transition-level scalar for each denoising step.

\begin{equation}
\ell_{\mathrm{trans}}
=
\sum_{l=1}^{L}\sum_{d=1}^{D_p}\hat{\ell}_{l,d}
\end{equation}
is replaced by a set of token-wise log-likelihoods
\begin{equation}
\ell_{\mathrm{token},l}
=
\sum_{d=1}^{D_p}\hat{\ell}_{l,d},
\qquad l \in \{1,\ldots,L\},
\end{equation}
where $D_p$ denotes the dimensionality of each latent patch.

This tokenization preserves the spatial variation of the policy likelihood across latent regions, providing a finer-grained optimization signal for subsequent spatial credit assignment.

\subsection{Spatial Gradient-Guided Credit Assignment}

To facilitate reproducibility, we provide the PyTorch-style implementation of our spatial gradient-guided credit assignment module. Given a differentiable reward function, we compute the reward gradient with respect to the generated latent representation and convert it into a normalized spatial credit map. The credit map is subsequently patchified to align with the $L$ latent tokens, thereby assigning spatially varying credit weights to the corresponding token-wise policy objectives.

\begin{lstlisting}[language=Python, caption={PyTorch-style implementation of spatial gradient-guided credit assignment.}, label={lst:spatial_credit}]
# PyTorch implementation
scorer = DifferentiableRewardFunction()

def _summed_reward_gradients(rewards, images):
    """Return image gradients using a single backward pass over summed rewards."""
    rewards = rewards.flatten()

    return torch.autograd.grad(
        rewards.sum(),
        images,
        create_graph=False,
    )[0]

def _grad_postprocess(gradient, temperature: float = 0.1):
    gradient = gradient.detach().float()

    gradient_magnitude = torch.linalg.vector_norm(
        gradient, ord=2, dim=1
    ).detach()
    eps = torch.finfo(torch.float32).eps
    flat = gradient_magnitude.flatten(1)

    median = torch.median(
        flat, dim=1
    ).values

    deviation = torch.abs(
        gradient_magnitude - median[:, None, None]
    )

    mad = torch.median(
        deviation.flatten(1), dim=1
    ).values

    normalized_gradient = (
        gradient_magnitude - median[:, None, None]
    ) / (
        mad[:, None, None] + eps
    )

    spatial_credits = torch.sigmoid(
        -normalized_gradient / temperature
    )
    return spatial_credits

def spatial_credit_assignment(images, prompts):
    images = images.detach().clone().requires_grad_(True)
    rewards = scorer(images, prompts)
    image_gradients = _summed_reward_gradients(rewards, images)
    spatial_credits = _grad_postprocess(image_gradients, temperature)
    
    return rewards.detach().clone(), spatial_credits

\end{lstlisting}

\section{Additional Experiment Details}

\paragraph{Ablation study on the clipping range.}

After tokenizing the transition-level log-likelihood, we observe that the clipping behavior changes noticeably compared with the original formulation. In conventional GRPO-style optimization, the clipping range is commonly set to $\epsilon=0.2$, which typically maintains a moderate clipping ratio and prevents excessively large policy updates. This choice provides a practical balance between update stability and effective policy improvement.

However, after decomposing the transition-level likelihood into token-wise components, the importance ratios are evaluated at a much finer granularity. Consequently, the distribution of token-wise ratios becomes different from that of the original transition-level ratio, making the conventional clipping range not necessarily optimal. We therefore investigate the effect of different clipping ranges under the tokenized likelihood formulation.

\begin{figure*}[t]
\centering
\includegraphics[width=1.0\linewidth,keepaspectratio]{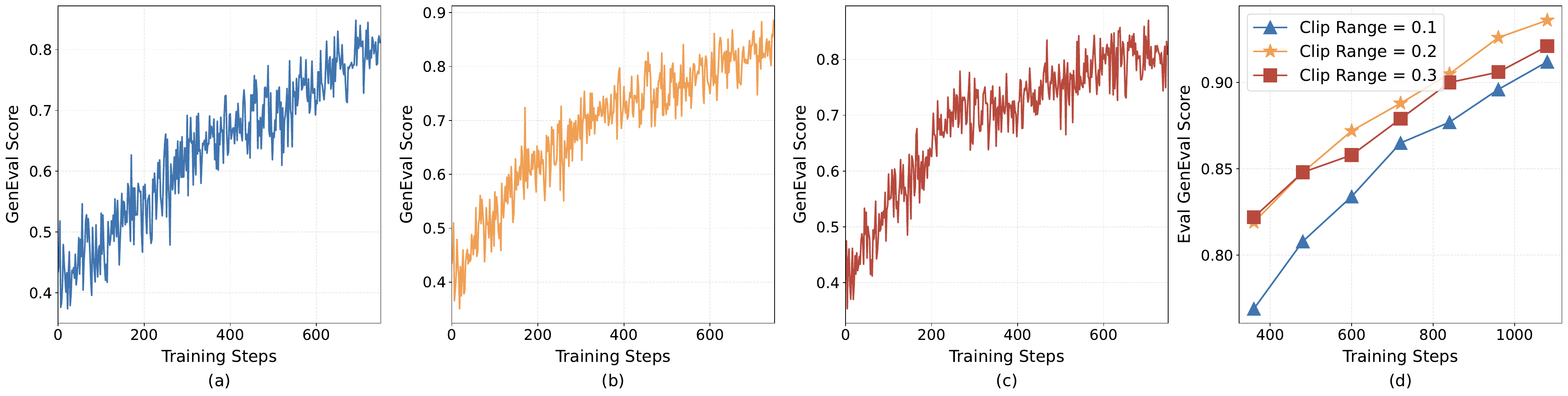}
\vspace{-4mm}
\caption{Ablation study on the clipping range $\epsilon$. Subfigures (a), (b), and (c) show the training reward curves with clipping ranges $\epsilon=0.1$, $0.2$, and $0.3$, respectively. Subfigure (d) compares the corresponding GenEval scores on the test set.}
 \label{fig:clip_range}
\end{figure*}

As shown in Fig.~\ref{fig:clip_range}, we compare $\epsilon=0.1$, $0.2$, and $0.3$. Different clipping ranges do not lead to a substantial difference in the overall convergence speed during training, indicating that the optimization remains relatively stable across these settings. Nevertheless, the clipping range has a clear impact on the final model performance. In particular, $\epsilon=0.2$ achieves the best performance on the test set, while either a more restrictive range ($\epsilon=0.1$) or a looser range ($\epsilon=0.3$) results in inferior generalization performance. These results suggest that a moderate clipping range remains preferable after likelihood tokenization, as it provides a suitable balance between constraining unstable token-wise updates and preserving sufficiently informative policy updates.

\paragraph{Overhead of Spatial Credit Map Generation.}

\begin{wraptable}{r}{0.44\textwidth}
\vspace{-4mm}
\centering
\caption{Runtime comparison of reward computation. All results are reported in seconds on an NVIDIA A100 80GB GPU.}
\vspace{5mm}
\label{tab:runtime_comparison}
\setlength{\tabcolsep}{3.5pt}
\renewcommand{\arraystretch}{1.05}
\resizebox{0.40\textwidth}{!}{
\begin{tabular}{c|cc|cc}
\toprule
\multirow{2}{*}{\textbf{BS}} & \multicolumn{2}{c|}{\textbf{PickScore}} & \multicolumn{2}{c}{\textbf{ImageReward}} \\
\cmidrule(lr){2-3}\cmidrule(lr){4-5}
& \textbf{Ori.} & \textbf{Diff.}
& \textbf{Ori.} & \textbf{Diff.} \\
\midrule
1  & 0.024  & 0.044  & 0.019  & 0.047  \\
4  & 0.059 & 0.109 & 0.023 & 0.059  \\
8  & 0.093 & 0.188 & 0.042  & 0.081  \\
16 & 0.162 & 0.343 & 0.075  & 0.158 \\
\bottomrule
\end{tabular}
}
\vspace{-3mm}
\end{wraptable}

We further analyze the additional computational overhead introduced by the spatial credit assignment module. Since log-likelihood patchification only involves lightweight tensor reshaping and local reduction operations, its computational cost is negligible. We therefore focus on the generation of the spatial credit map, which constitutes the main additional computation in our method.

Specifically, constructing the spatial credit map requires computing the gradient of a differentiable reward with respect to the generated representation, followed by lightweight post-processing operations including normalization and patchification. To quantify this overhead, we measure the wall-clock runtime of the original and differentiable implementations of PickScore and ImageReward under different batch sizes on an NVIDIA A100 80GB GPU.

As shown in Table~\ref{tab:runtime_comparison}, enabling gradient computation consistently increases the reward-evaluation time, with the differentiable implementations requiring approximately $1.8\times$--$2.5\times$ the runtime of their original counterparts across the evaluated batch sizes. For instance, at a batch size of 16, the runtime of PickScore increases from $0.162$\,ms to $0.343$\,s, while that of ImageReward increases from $0.075$\,ms to $0.158$\,ms. Nevertheless, the absolute runtime remains below $0.35$\,s even at the largest evaluated batch size. These results indicate that the spatial credit map can be obtained with a limited computational overhead, while enabling fine-grained spatial credit assignment for token-wise policy optimization.

\paragraph{More Results on Pick-a-Pic.}

\begin{figure*}[h]
\centering
\includegraphics[width=1.0\linewidth,keepaspectratio]{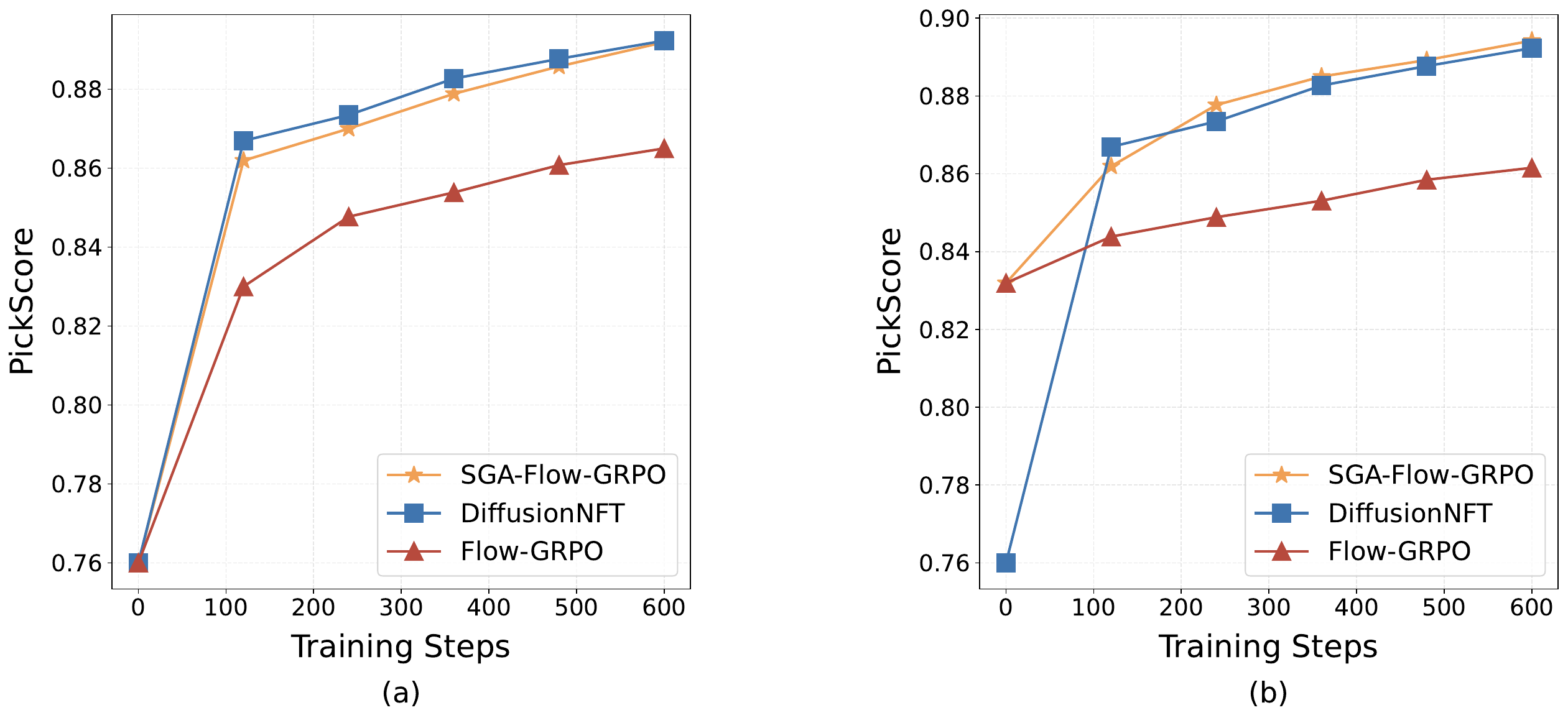}
\vspace{-5mm}
\caption{PickScore comparison on Pick-a-Pic.
We compare our method with DiffusionNFT and Flow-GRPO under $\mathrm{CFG}=1.0$ and $\mathrm{CFG}=4.5$. All methods are trained for 600 steps, and test-set PickScore is evaluated every 120 training steps.
}
\label{fig:comp_pickscore}
\end{figure*}

We further evaluate our method on the Pick-a-Pic~\citep{pickscore_2023} dataset using PickScore as the evaluation metric. We compare our method with DiffusionNFT and Flow-GRPO under two classifier-free guidance settings, i.e., $\mathrm{CFG}=1.0$ and $\mathrm{CFG}=4.5$. All methods are optimized for 600 training steps, and we evaluate the corresponding checkpoints on the test set every 120 steps.

As shown in Fig.~\ref{fig:comp_pickscore}, our method consistently achieves stronger PickScore performance throughout training under both CFG settings. Under $\mathrm{CFG}=1.0$, our method exhibits faster improvement and maintains a clear advantage over the compared baselines as training progresses. A similar trend is observed under $\mathrm{CFG}=4.5$, demonstrating that the proposed spatially fine-grained credit assignment remains effective under stronger classifier-free guidance. These results further indicate that our method provides more effective policy updates and generalizes well beyond the GenEval benchmark.

We further present qualitative results on the Pick-a-Pic dataset. We compare our method with DiffusionNFT and Flow-GRPO under $\mathrm{CFG}=1.0$ and $\mathrm{CFG}=4.5$. 

\begin{figure*}[h]
\centering
\includegraphics[width=1.0\linewidth,keepaspectratio]{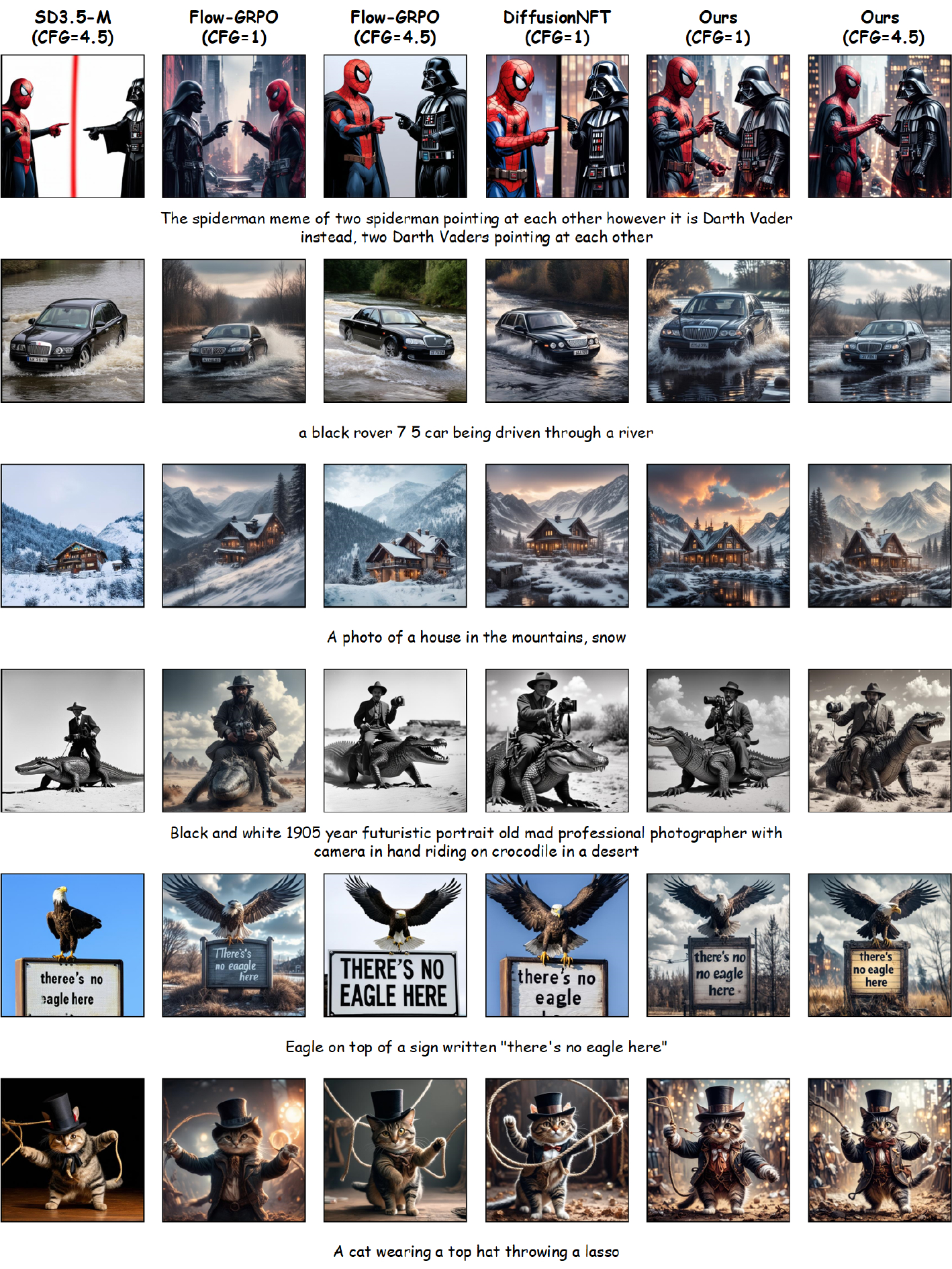}
\vspace{-5mm}
\caption{Qualitative Comparison on Pick-a-Pic.
}
\label{fig:qual_pickapic}
\end{figure*}

As shown in Fig.~\ref{fig:qual_pickapic}, our method produces images with better semantic alignment and overall visual quality. In particular, compared with the baselines, our method generates results that are more consistent with the text prompts while preserving more coherent object structures and richer visual details.

\end{document}